\documentclass[english]{esannV2}
\usepackage[T1]{fontenc}
\usepackage[latin9]{inputenc}
\usepackage{array}
\usepackage{rotating}
\usepackage{multirow}
\usepackage{amsbsy}
\usepackage{amsmath}
\usepackage{amssymb}
\usepackage{esint}
\usepackage{amstext}
\usepackage[bf]{caption2}

\makeatletter

\usepackage{array}

\makeatother
\usepackage{babel}
\begin{document}
%style file for ESANN manuscripts

\title{Regularization in Relevance Learning Vector Quantization Using $l_{1}$-Norms}

%***********************************************************************
% AUTHORS INFORMATION AREA
%***********************************************************************

\author{M. Riedel$^{1}$\footnote{M.R. and M.K. are supported by a grant of the ESF, Saxony, Germany.}, F. Rossi$^{2}$, M. K\"{a}stner$^{1*}$, and T. Villmann$^{1}$ % Optional short acknowledgment: remove next line if non-needed
%\thanks{This is an optional funding source acknowledgement.}
% DO NOT MODIFY THE FOLLOWING '\vspace' ARGUMENT
\vspace{0.3cm}
\\
 % Addresses and institutions (remove "1- " in case of a single institution)
1- University of Appl. Sciences Mittweida - Dept. of Mathematics \\
 Mittweida, Saxonia - Germany % Remove the next three lines in case of a single institution
\vspace{0.1cm}
\\
 2- University Paris Sorbonne-Pantheon, France\\
 }

%***********************************************************************
% END OF AUTHORS INFORMATION AREA
%***********************************************************************

\vspace{-0.3cm}
\date{}
\maketitle

\begin{abstract}
We propose in this contribution a method for $l_{1}$-regularization
in prototype based relevance learning vector quantization (LVQ) for
sparse relevance profiles. Sparse relevance profiles in hyperspectral
data analysis fade down those spectral bands which are not necessary
for classification. In particular, we consider the sparsity in the
relevance profile enforced by LASSO optimization. The latter one is
obtained by a gradient learning scheme using a differentiable parametrized
approximation of the $l_{1}$-norm, which has an upper error bound.
We extend this regularization idea also to the matrix learning variant
of LVQ as the natural generalization of relevance learning. 
\end{abstract}

\section{Introduction}

Learning vector quantization (LVQ) as proposed by T. Kohonen
is one of the most popular methods for prototype based classification
of vectorized data \cite{kohonen95a}. Sato\&Yamada proposed
a modification of this approach such that the learning heuristic of
LVQ is replaced by a stochastic gradient descent learning based on
a cost function \cite{sato96a}. The cost function is an approximation
of the usual classification error based on dissimilarity evaluations
for the best matching prototypes. This generalized LVQ (GLVQ) optimizes
the hypothesis margin \cite{Crammer2002a}. An improvement of GLVQ
performance can be obtained by relevance learning (GRLVQ), i.e. weighting
the data dimensions to distinguish the data classes \cite{Villmann2002d}.
High weighting values indicate high relevance. The resulting relevance
profile provides the information about the importance of the data
dimensions for the classification to be learned. Frequently, small
but non-vanishing relevance values are obtained for large parts of
the relevance profiles. This problem frequently occurs for high-dimensional
data like hyperspectra. This behavior is not sufficient in the light
of \emph{sparse models}, where negligible spectral bands should be
dropped off, if the classification accuracy is sufficiently high.

In this contribution we propose a $l_{1}$-regularization approach
to obtain sparsity in relevance learning, i.e. sparsity in the relevance
profile \cite{Villmann2012b}. It is based on the \textbf{L}east \textbf{A}bsolute
\textbf{S}election and \textbf{S}hrinkage \textbf{O}perator approach
(LASSO, \cite{TibshiraniLASSO1996}) but realizing a gradient descent
learning scheme whereas original LASSO uses convex optimization. For this purpose, a \emph{differentiable approximation}
of the $l_{1}$-norm is considered \cite{Schmidt2007}. We show further
that this approach can easily be transferred to the matrix learning
GLVQ (GMLVQ, \cite{Schneider2009_MatrixLearning}) using the consistent
matrix norm. We illustrate the method for classification coffee hyperspectra
to distinguish different coffee sorts.

\section{Generalized Relevance and Matrix LVQ}

We suppose for learning vector quantization approaches that the data
are given as vectors $\mathbf{v}\in V\subseteq\mathbb{R}^{n}$, and
the prototypes of the LVQ model are collected in the set $W=\left\{ \mathbf{w}_{k}\in\mathbb{R}^{n},k=1\ldots M\right\} $.
Each training data vector $\mathbf{v}$ belongs to a class $x_{\mathbf{v}}\in\mathcal{C}=\left\{ 1,\ldots,C\right\} $.
The prototypes have labels $y_{\mathbf{w}_{k}}\in\mathcal{C}$ indicating
their responsibility to the several classes. The GLVQ approach approximates
the classification error to be minimized by the cost function 
\begin{equation}
E\left(W\right)=\frac{1}{2}\sum_{\mathbf{v}\in V}f\left(\mu\left(\mathbf{v}\right)\right)\textrm{ with }\mu\left(\mathbf{v}\right)=\frac{d^{+}\left(\mathbf{v}\right)-d^{-}\left(\mathbf{v}\right)}{d^{+}\left(\mathbf{v}\right)+d^{-}\left(\mathbf{v}\right)}\label{cost_function_GLVQ}
\end{equation}
as the classifier function and $d^{+}\left(\mathbf{v}\right)=d\left(\mathbf{v},\mathbf{w}^{+}\right)$
denotes the dissimilarity between the data vector $\mathbf{v}$ and
the closest prototype $\mathbf{w}^{+}$ with the same class label
$y_{\mathbf{w}^{+}}=x_{v}$, and $d^{-}\left(\mathbf{v}\right)=d\left(\mathbf{v},\mathbf{w}^{-}\right)$
is the dissimilarity degree for the best matching prototype $\mathbf{w}^{-}$
with a class label $y_{\mathbf{w}^{-}}$ different from $x_{\mathbf{v}}$.
The classifier function $\mu\left(\mathbf{v}\right)$ becomes negative
if the data point is classified correctly. The transformation function
$f$ is a monotonically increasing function usually chosen as sigmoid
or the identity function. The dissimilarity measure $d\left(\mathbf{v},\mathbf{w}\right)$
is usually chosen as the squared Euclidean distance.

Learning in GLVQ of $\mathbf{w}^{+}$ and $\mathbf{w}^{-}$ is performed
by the \emph{stochastic} gradient with respect to the cost function
$E\left(W\right)$ for a given data vector $\mathbf{v}$ according
to 
\begin{equation}
\frac{\partial_{S}E\left(W\right)}{\partial\mathbf{w}^{+}}=\xi^{+}\cdot\frac{\partial d^{+}}{\partial\mathbf{w}^{+}}\text{ and }\frac{\partial_{S}E\left(W\right)}{\partial\mathbf{w}^{-}}=\xi^{-}\cdot\frac{\partial d^{-}}{\partial\mathbf{w}^{-}}\label{eq:prototype_update}
\end{equation}
with 
\begin{equation}
\xi^{+}=f^{\prime}\cdot\frac{2\cdot d^{-}\left(\mathbf{v}\right)}{\left(d^{+}\left(\mathbf{v}\right)+d^{-}\left(\mathbf{v}\right)\right)^{2}}\text{ and }\xi^{-}=-f^{\prime}\cdot\frac{2\cdot d^{+}\left(\mathbf{v}\right)}{\left(d^{+}\left(\mathbf{v}\right)+d^{-}\left(\mathbf{v}\right)\right)^{2}}.\label{eq:xi}
\end{equation}
For the squared Euclidean metric we simply have the derivative $\frac{\partial d^{\pm}\left(\mathbf{v}\right)}{\partial\mathbf{w}^{\pm}}=-2\left(\mathbf{v}-\mathbf{w}^{\pm}\right)$
realizing a vector shift of the prototypes.

Standard relevance learning replaces the squared Euclidean distance
in GLVQ by a parametrized bilinear form 
\begin{equation}
d_{\Lambda}\left(\mathbf{v},\mathbf{w}\right)=\left(\mathbf{v}-\mathbf{w}\right)^{\top}\Lambda\left(\mathbf{v}-\mathbf{w}\right)\label{eq:relevance metric}
\end{equation}
with $\Lambda$ being a positive semi-definite \emph{diagonal matrix}
\cite{Villmann2002d}. The diagonal elements $\lambda_{i}=\sqrt{\Lambda_{ii}}$
form the relevance profile weighting the data dimensions. During the
learning phase, the relevance parameter $\lambda_{i}$ are adapted
according to 
\begin{equation}
\triangle\mathbf{\Lambda}\sim-\frac{\partial_{S}E\left(W\right)}{\partial\mathbf{\Lambda}}=-\xi^{+}\cdot\frac{\partial d_{\Lambda}^{+}\left(\mathbf{v}\right)}{\partial\lambda_{j}}-\xi^{-}\cdot\frac{\partial d_{\Lambda}^{-}\left(\mathbf{v}\right)}{\partial\lambda_{j}}\label{eq:relevance_update}
\end{equation}
realizing a stochastic gradient descent. An subsequent normalization
has to be applied such that $\sum_{i}\lambda_{i}^{2}=\sum_{i}\Lambda_{i,i}=1$
is assured. 

The obvious generalization of this scheme is to take the matrix $\Lambda$
as a positive semi-definite quadratic form $\Lambda=\Omega^{\top}\Omega$
with an arbitrary matrix $\Omega\in\mathbb{R}^{m\times n}$ \cite{Villmann2012a,Schneider2009_MatrixLearning}.
To avoid degeneracy $\det\left(\Lambda\right)>0$ is required \cite{Villmann2010l}.
Then equation (\ref{eq:relevance metric}) can be written as $d_{\Omega}\left(\mathbf{v},\mathbf{w}\right)=\left(\Omega\left(\mathbf{v}-\mathbf{w}\right)\right)^{2}$.
The resulting derivatives in (\ref{eq:prototype_update}) are obtained
as $\frac{\partial d^{\pm}\left(\mathbf{v}\right)}{\partial\mathbf{w}^{\pm}}=-2\Lambda\left(\mathbf{v}-\mathbf{w}^{\pm}\right)$,
which are accompanied by the $\Omega$-update 
\begin{equation}
\triangle\Omega_{r_{1},r_{2}}\sim-\frac{\partial_{S}E\left(W\right)}{\partial\Omega_{r_{1},r_{2}}}=\xi^{+}\cdot\frac{\partial d_{\Omega}^{+}\left(\mathbf{v}\right)}{\partial\Omega_{r_{1},r_{2}}}+\xi^{-}\cdot\frac{\partial d_{\Omega}^{-}\left(\mathbf{v}\right)}{\partial\Omega_{r_{1},r_{2}}}\label{eq:Omega_update}
\end{equation}
and subsequent normalization $\sum_{i,j}\Omega_{i,j}^{2}=1$ \cite{Villmann2010l}.
We refer to this matrix variant as GMLVQ.

\section{Sparsity in Relevance and Matrix Learning by Gradient LASSO Learning }

For the LASSO method it is
assumed that we want to optimize a cost function depending on a parameter
vector $\mathbf{\lambda}$ which has to follow a regularization condition
according to the $l_{1}$-norm \cite{HastieTibshiraniBook2001,TibshiraniLASSO1996}. In the context of GRLVQ this cost
function is $E\left(W,\mathbf{\lambda}\right)$ according to (\ref{cost_function_GLVQ})
with the parameters $\lambda_{i}$ obtained from the relevance metric
(\ref{eq:relevance metric}). The LASSO approach adds a regularization
term such that 
\begin{equation}
\min\limits _{\mathbf{\lambda}}E^{*}\left(W,\mathbf{\lambda}\right)=E\left(W,\mathbf{\lambda}\right)+\xi\left\Vert \mathbf{\lambda}\right\Vert _{1}\label{eq:LASSO model}
\end{equation}
with a weighting factor $\xi>0$. Many optimization methods are known
to solve this problem. Yet, in the context of gradient descent learning
in GRLVQ it would be desirable to have a gradient learning scheme
of LASSO, too. However, the regularization term $R\left(\mathbf{\lambda}\right)=\left\Vert \mathbf{\lambda}\right\Vert _{1}=\sum\limits _{i=1}^{n}\vert\lambda_{i}\vert$
is not differentiable with respect to the $\lambda_{i}$. Fortunately,
a differentiable approximation for $R\left(\mathbf{\lambda}\right)$
can be found \cite{Schmidt2007}: We split the absolute value $\vert x\vert$
into $\vert x\vert=\left(x\right)_{+}+\left(-x\right)_{+}$ with $\left(x\right)_{+}=\max\left\lbrace x,0\right\rbrace $.
This allows an approximation $\vert x\vert_{\alpha}$ of $\vert x\vert$
using the relation 
\begin{equation}
\left(x\right)_{+}\approx x+\frac{1}{\alpha}\ln\left(1+e^{-\alpha x}\right)
\end{equation}
depending on the approximation parameter $\alpha$ \cite{ChengMangasarian}.
We obtain 
\begin{equation}
\vert x\vert_{\alpha}=\frac{1}{\alpha}\ln\left(2+e^{-\alpha x}+e^{\alpha x}\right)\label{eq:l1 norm approximation}
\end{equation}
with the upper bound $\vert\vert x\vert-\vert x\vert_{\alpha}\vert\leq2\frac{\ln2}{\alpha}$.
Inserting this in $R\left(\mathbf{\lambda}\right)$ the gradients
are obtained as
\begin{equation}
\frac{\partial R\left(\mathbf{\lambda}\right)}{\partial\lambda_{j}}\approx\tanh\left(\frac{\alpha\lambda_{j}}{2}\right).
\end{equation}

Analogously, for GMLVQ with $l_{1}$-regularization via LASSO a regularization
term $R\left(\Omega\right)=\left\Vert \Omega\right\Vert _{1}$ is added with
\begin{equation}
\Vert\Omega\Vert_{1}=\max\limits _{1\leq j\leq n}\sum\limits _{i=1}^{m}\vert\Omega_{ij}\vert
\end{equation}
being the matrix norm consistent to the $l_{1}$-vector norm. Using
the recursion $\max\left(x_{1},x_{2},\ldots,x_{n}\right)=\max\left(x_{1},\max\left(x_{2},\ldots,x_{n}\right)\right)$
and the relation $\max\left(x,y\right)=\frac{1}{2}\left(x+y-\vert x-y\vert\right)$
the regularization term dependency becomes $R\left(\Omega\right)=R\left(\vert\Omega_{ij}\vert\right)$.
Thus we can apply again the above approximation (\ref{eq:l1 norm approximation}).
A lengthy but simple calculation yields 
\begin{equation}
\frac{\partial R\left(\Omega\right)}{\partial\Omega_{st}}\approx\frac{1}{2}\tanh\left(\frac{\alpha\Omega_{st}}{2}\right)-\frac{T}{2}
\end{equation}
with
\[
T=\frac{\exp\left(-\alpha\left(\Omega_{st}+\overline{\Omega}_{st}\right)\right)\cdot\left(\exp\left(2\alpha\Omega_{st}\right)-1\right)\cdot\left(\exp\left(2\alpha\overline{\Omega}_{st}\right)-\frac{\exp\left(2\alpha\Omega_{st}\right)}{\left(1+\exp\left(\alpha\Omega_{st}\right)\right)^{4}}\right)}{2+\exp\left(-\alpha\left(\Omega_{st}-\overline{\Omega}_{st}\right)\right)+\exp\left(\alpha\left(\Omega_{st}+\overline{\Omega}_{st}\right)\right)+\frac{\exp\left(\alpha\left(\Omega_{st}-\overline{\Omega}_{st}\right)\right)}{\left(1+\exp\left(\alpha\Omega_{st}\right)\right)^{2}}}
\]
and
\begin{equation}
\overline{\Omega}_{st}=\sum\limits _{i=1;i\neq s}^{m}\vert\Omega_{it}\vert_{\alpha}-\max\limits _{1\leq j\leq d;j\neq t}\sum\limits _{i=1}^{m}\vert\Omega_{ij}\vert_{\alpha}
\end{equation}
Further, it can be shown that $\frac{1}{m}\Vert\Omega\Vert_{1}^{2}\leq\Vert\Lambda\Vert_{1}\leq n\Vert\Omega\Vert_{1}^{2}$
is valid.

In conclusion, we derived a differentiable approximation of the $l_{1}$-regularization which can be used in gradient descent learning of, for example, GRLVQ and GMLVQ. 

\section{Simulation Results}

\begin{figure}[htp]
\centerline{\includegraphics[width=0.7\linewidth]{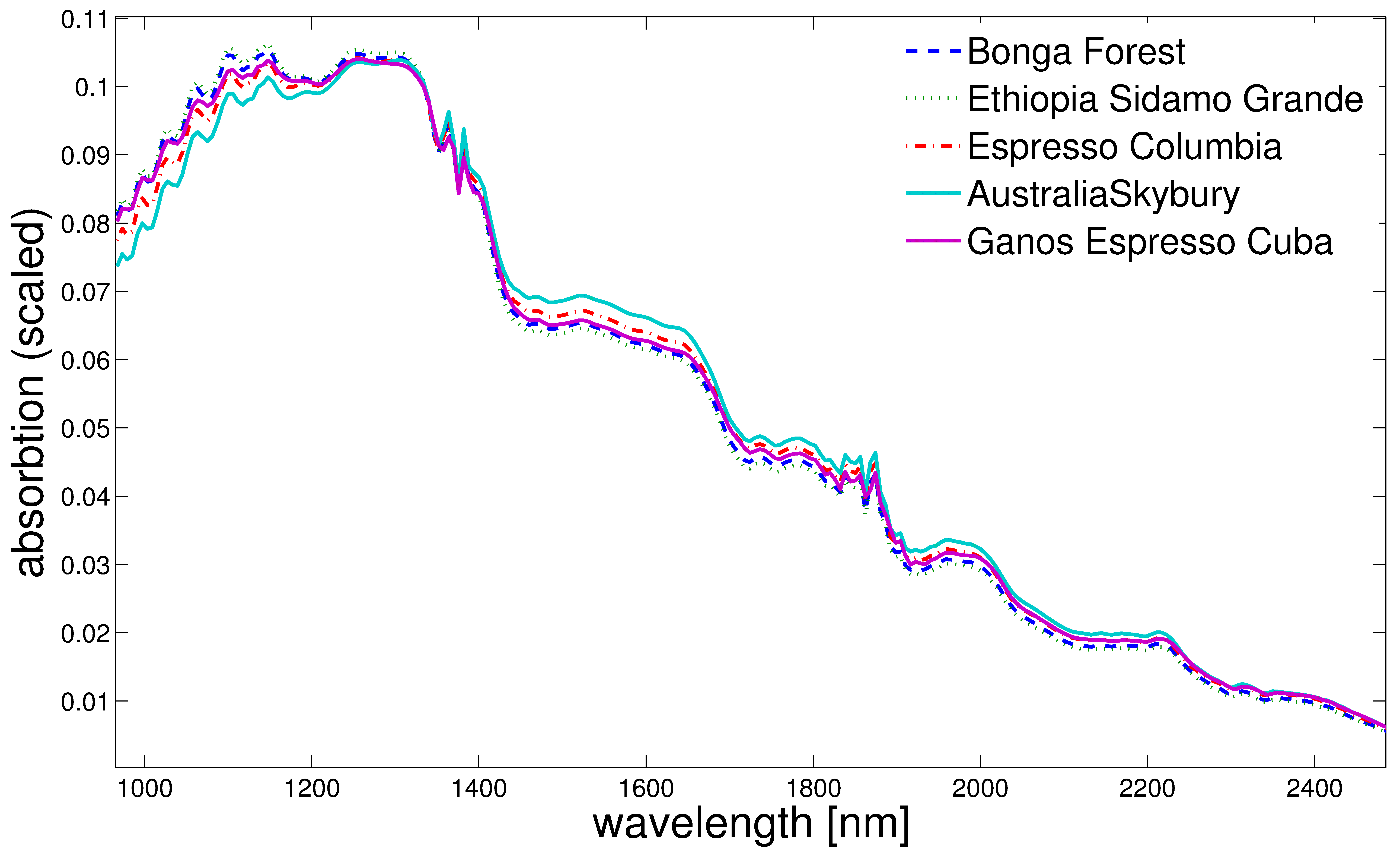}}
\caption{Mean spectra of the five investigated coffee types.}

\label{fig:mean_spectra_treated} 
\end{figure}

We applied the sparsity relevance learning model to classify hyperspectral
short-wave infrared range (SWIR) spectral vectors of five coffee types.
Hyperspectral processing along with an appropriate analysis of the
acquired high-dimensional spectra has proven to be a suitable and
very powerful method to quantitatively assess the biochemical composition
of a wide range of biological samples \cite{BackFebollSff_Whispers11}.
By utilizing a hyperspectral camera (HySpex SWIR-$320$m-e, Norsk
Elektro Optikk A/S) we obtained a rather extensive data base of spectra
of five different coffee types ($5000$ spectra for each class). We
used spectra in the SWIR between $970$ nm and $2,500$ nm at $6$
nm resolution yielding $256$ bands per spectrum. Proper image calibration
was done by using a standard reflection pad (polytetrafluoroethylene,
PTFE)\cite{BackFebollSff_CoCoTea11}. After appropriate image segmentation
the obtained spectra were normalized according to the $l_{2}$-norm
and reduced to $200$ bands ignoring the range $2,000-2,500$ nm.
The mean spectra of the five types are visualized in Fig. \ref{fig:mean_spectra_treated}. 

\begin{figure}[htp]
\centerline{\includegraphics[width=0.7\linewidth]{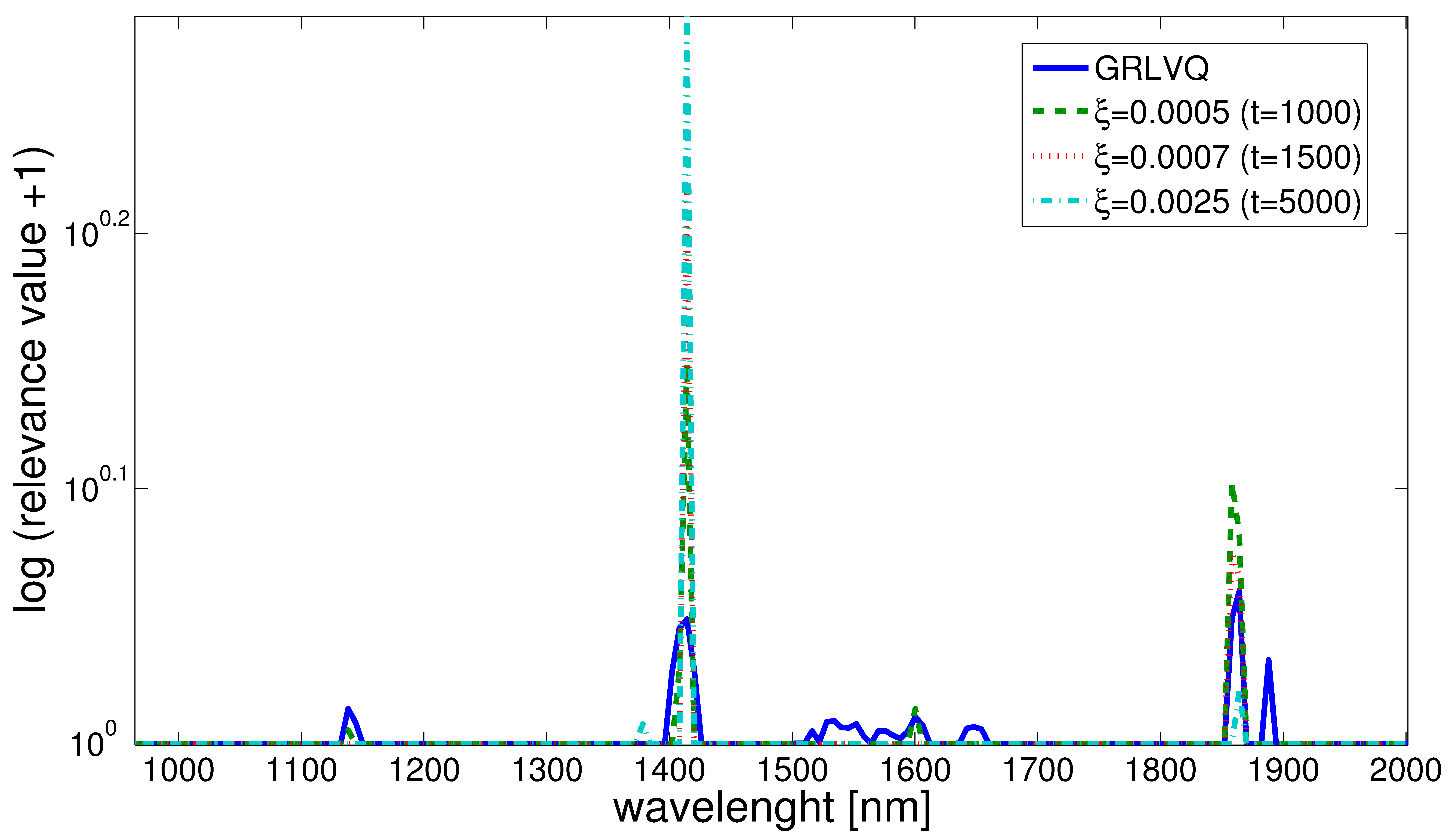}}
\caption{Development of the sparsity of the relevance profile during LASSO-learning.
With increasing influence of the regularization term the profile becomes
sparse.}

\label{fig:relevance profiles} 
\end{figure}

\begin{figure}[htp]
\centerline{\includegraphics[width=0.7\linewidth]{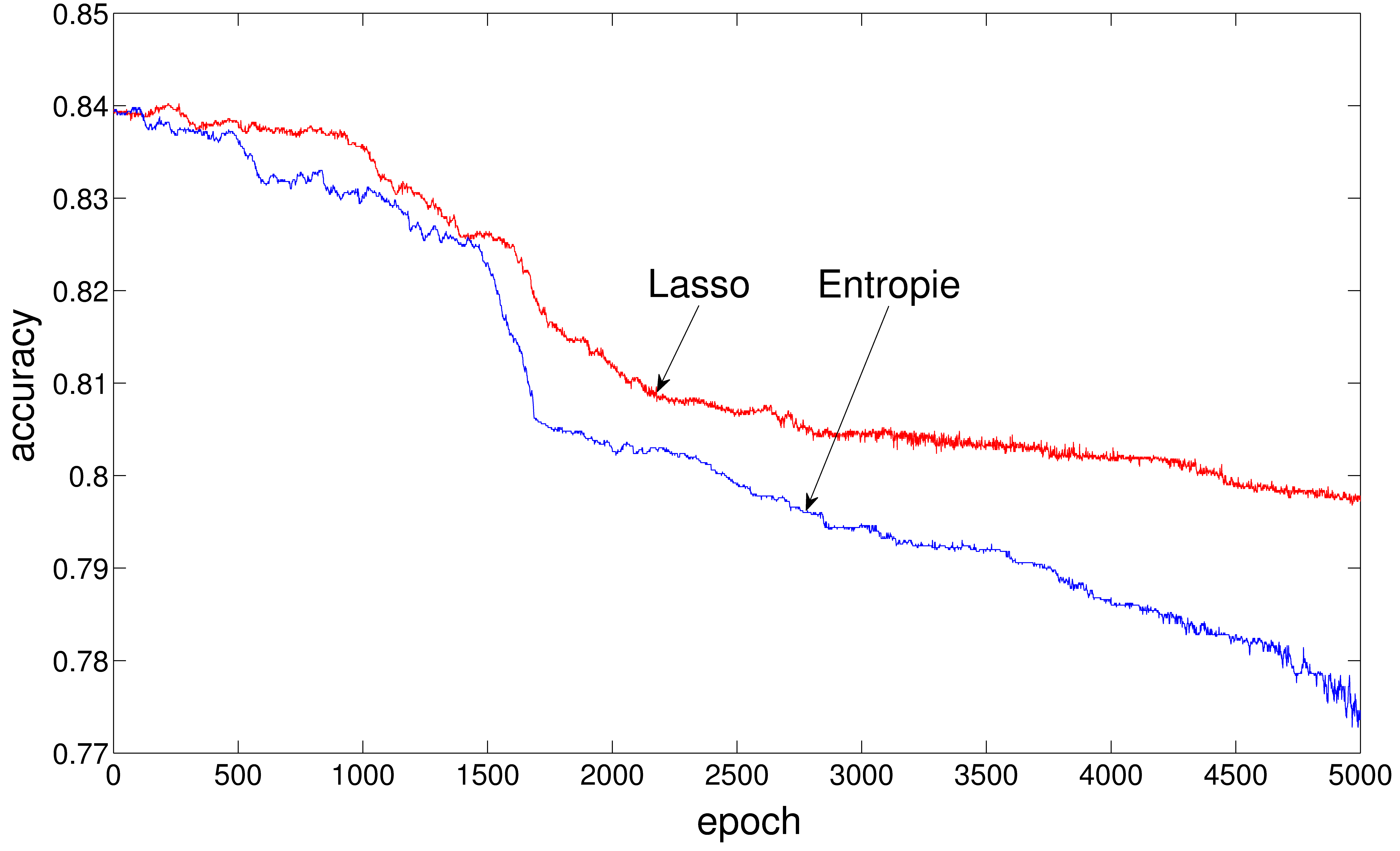}}
\caption{Development of the accuracies during sparsity adaptation according
LASSO (red) and entropy based (blue) regularization. We observe instabilities
of the entropy based method in the final phase of regularization.}

\label{fig:accuracies} 
\end{figure}

After standard training the GRLVQ model with full relevance profile
yields $83,96\%$ accuracy. Starting with this solution the LASSO-model
(\ref{eq:LASSO model}) was applied with linearly increasing weighting
factor $\xi$ of the regularization term, the approximation parameter
$\alpha$ in (\ref{eq:l1 norm approximation}) was set constant $\alpha=5$.
We compare this LASSO-approach with a sparsity model based on an entropy
penalty term added to the cost function of GRLVQ as suggested in \cite{Villmann2012b}.
Both models enforce the sparsity of the relevance profiles. We depict
the results of the LASSO approach Fig. \ref{fig:relevance profiles},
the other result is similar and has dropped because the lack of space.
However, the accuracy decrease differs. LASSO keeps longer a
high accuracy than the entropy approach, see Fig. \ref{fig:accuracies} .
 Moreover, the entropy based method shows heavy instabilities
if the relevance weights for spectral bands approach zero values at
the end of the regularization process.

\section{Conclusion}

Sparsity in hyperspectral data analysis play an important role to
concentrate on those bands, which are important for classification.
Relevance learning as proposed in GRLVQ offers a possibility to weight
the bands. However, frequently it delivers small but non-vanishing
weights. Additional regularization can help to obtain sparse models.
We have shown in this contribution that LASSO $l_{1}$-regularization
can be applied in gradient based online learning using a differentiable
approximation. We illustrate the method for an exemplary application
of coffee classification based on hyperspectral signatures.

\bibliographystyle{abbrv}

\begin{thebibliography}{}

\end{thebibliography}


\begin{thebibliography}{10}

\bibitem{BackFebollSff_CoCoTea11}
A.~Backhaus, F.~Bollenbeck, and U.~Seiffert.
\newblock High-throughput quality control of coffee varieties and blends by
  artificial neural networks and hyperspectral imaging.
\newblock In {\em Proceedings of the 1st International Congress on Cocoa,
  Coffee and Tea, CoCoTea 2011}, page accepted for publication, 2011.

\bibitem{BackFebollSff_Whispers11}
A.~Backhaus, F.~Bollenbeck, and U.~Seiffert.
\newblock Robust classification of the nutrition state in crop plants by
  hyperspectral imaging and artificial neural networks.
\newblock In {\em Proceedings of the 3rd IEEE Workshop on Hyperspectral Imaging
  and Signal Processing: Evolution in Remote Sensing WHISPERS 2011}, page~9.
  IEEE Press, 2011.

\bibitem{Villmann2012a}
K.~Bunte, P.~Schneider, B.~Hammer, F.-M. Schleif, T.~Villmann, and M.~Biehl.
\newblock Limited rank matrix learning, discriminative dimension reduction and
  visualization.
\newblock {\em Neural Networks}, 26(1):159--173, 2012.

\bibitem{ChengMangasarian}
C.~Chen and O.~Mangasarian.
\newblock Smoothing methods for convex inequalities and linear complementarity
  problems.
\newblock {\em Mathematical Programming}, 71(1):51--69, 1995.

\bibitem{Crammer2002a}
K.~Crammer, R.~Gilad-Bachrach, A.Navot, and A.Tishby.
\newblock Margin analysis of the {LVQ} algorithm.
\newblock In S.~Becker, S.~Thrun, and K.~Obermayer, editors, {\em Advances in
  Neural Information Processing (Proc. NIPS 2002)}, volume~15, pages 462--469,
  Cambridge, MA, 2003. MIT Press.

\bibitem{Villmann2002d}
B.~Hammer and T.~Villmann.
\newblock Generalized relevance learning vector quantization.
\newblock {\em Neural Networks}, 15(8-9):1059--1068, 2002.

\bibitem{HastieTibshiraniBook2001}
T.~Hastie, R.~Tibshirani, and J.~Friedman.
\newblock {\em The Elements of Statistical Learning}.
\newblock Springer Verlag, Heidelberg-Berlin, 2001.

\bibitem{Villmann2012b}
M.~K\"{a}stner, B.~Hammer, M.~Biehl, and T.~Villmann.
\newblock Functional relevance learning in generalized learning vector
  quantization.
\newblock {\em Neurocomputing}, 90(9):85--95, 2012.

\bibitem{kohonen95a}
T.~Kohonen.
\newblock {\em {Self-Organizing Maps}}, volume~30 of {\em Springer Series in
  Information Sciences}.
\newblock Springer, Berlin, Heidelberg, 1995.
\newblock (Second Extended Edition 1997).

\bibitem{sato96a}
A.~Sato and K.~Yamada.
\newblock Generalized learning vector quantization.
\newblock In D.~S. Touretzky, M.~C. Mozer, and M.~E. Hasselmo, editors, {\em
  Advances in Neural Information Processing Systems 8. Proceedings of the 1995
  Conference}, pages 423--9. MIT Press, Cambridge, MA, USA, 1996.

\bibitem{Schmidt2007}
M.~Schmidt, G.~Fung, and R.~Rosales.
\newblock Fast optimization methods for l1 regularization: A comparative study
  and two new approaches.
\newblock In J.~Kok, J.~Koronacki, R.~Mantaras, S.~Matwin, D.~Mladeni\v{c}, and
  A.~Skowron, editors, {\em Machine Learning: ECML 2007}, volume 4701 of {\em
  Lecture Notes in Computer Science}, chapter~28, pages 286--297. Springer
  Berlin Heidelberg, Berlin, Heidelberg, 2007.

\bibitem{Villmann2010l}
P.~Schneider, K.~Bunte, H.~Stiekema, B.~Hammer, T.~Villmann, and M.~Biehl.
\newblock Regularization in matrix relevance learning.
\newblock {\em IEEE Transactions on Neural Networks}, 21(5):831--840, 2010.

\bibitem{Schneider2009_MatrixLearning}
P.~Schneider, B.~Hammer, and M.~Biehl.
\newblock Adaptive relevance matrices in learning vector quantization.
\newblock {\em Neural Computation}, 21:3532--3561, 2009.

\bibitem{TibshiraniLASSO1996}
R.~Tibshirani.
\newblock Regression shrinkage and selection via the lasso.
\newblock {\em Journal of the Royal Statistical Society: Series B},
  58(1):267--288, 1996.

\end{thebibliography}
\begin{footnotesize}

\end{footnotesize}

\end{document}